\documentclass[10pt,letterpaper,twocolumn]{article}
\usepackage[latin9]{inputenc}

\usepackage{color}
\usepackage{array}
\usepackage{units}
\usepackage{multirow}
\usepackage{amsmath}
\usepackage{amssymb}
\usepackage{graphicx}
\usepackage[unicode=true,
 bookmarks=false,
 breaklinks=true,pdfborder={0 0 1},backref=section,colorlinks=false]
 {hyperref}
\hypersetup{
 pagebackref=true,letterpaper=true,colorlinks}

\makeatletter

\pdfpageheight\paperheight
\pdfpagewidth\paperwidth

\providecommand{\tabularnewline}{\\}

\usepackage{iccv}
\usepackage{times}
\usepackage{epsfig}

\usepackage{subfigure}
\usepackage{multirow}
\usepackage{array}
\usepackage{textcomp}

\usepackage{nicefrac}

\iccvfinalcopy

\def\iccvPaperID{2788} 

\ificcvfinal\fi

\begin{document}

\title{Multi-scale Deep Learning Architectures for Person Re-identification}

\author{Xuelin Qian$^{1}$ ~Yanwei Fu$^{2,5,}$\thanks{Corresponding Author} ~ Yu-Gang Jiang$^{1,3}$ ~Tao Xiang$^4$ ~Xiangyang Xue$^{1,2}$\\
$^1$Shanghai Key Lab of Intelligent Info. Processing, School of Computer Science, Fudan University; \\
$^2$School of Data Science, Fudan University; $^3$Tencent AI Lab;\\
$^4$Queen Mary University of London; $^5$University of Technology Sydney;\\
{\tt\small \{15110240002,yanweifu,ygj,xyxue\}@fudan.edu.cn; t.xiang@qmul.ac.uk}  \\
}

\maketitle
\thispagestyle{empty}

\begin{abstract}
Person Re-identification (re-id) aims to match people across non-overlapping camera views in a public space. It is a challenging problem because many people captured in surveillance videos wear similar clothes. Consequently, the differences in their appearance are often subtle and only detectable at the right location and scales. Existing re-id models, particularly the recently proposed deep learning based ones match people at a single scale. In contrast, in this paper, a novel multi-scale deep learning model is proposed. Our model is able to
learn deep discriminative feature representations at different scales and automatically determine the most suitable scales for matching. The importance of different spatial locations for extracting discriminative features is also learned explicitly. Experiments are carried out to demonstrate that the proposed model outperforms the state-of-the art on a number of benchmarks. 
\end{abstract}

\section{Introduction}

Person re-identification (re-id) is defined as the task of matching
two pedestrian images crossing non-overlapping camera views \cite{gong2014person}.
It plays an important role in a number of applications in video surveillance,
including multi-camera tracking \cite{berclaz2008behavioral_maps,mensink2007distributedem},
crowd counting \cite{chan2009counting,ge2009crowd_count}, and multi-camera
activity analysis \cite{wang2008correspondencefree,wang2008npbayes}.
Person re-id is extremely challenging and remains unsolved for a number
of reasons. First, in different camera views, one person's appearance
often changes dramatically caused by the variances in body pose, camera
viewpoints, occlusion and illumination conditions. Second, in a public
space, many people often wear very similar clothes (e.g., dark coats
in winter). The differences that can be used to tell them apart are often subtle,
which could be the global, e.g., one person is bulkier than the
other, or local, e.g., the two people wear different shoes.

Early re-id methods use hand-crafted features for person appearance
representation and employ distance metric learning models as matching
functions. They focus on either  designing cross-view robust features \cite{farenzena2010reidentify_symmetry,viewpoint,Rivlin_tpami2013,mid_rui_zhao,XQDA},
or learning robust distance metrics \cite{LADF,metric_tpami14,seanPAMIpreReID,KISSME,kiss_metric,person_reid_salience,MtMCML,XQDA}, 
or both  \cite{larry_davis,XQDA,MFA,zhang_cooccurrence}.
Recently, inspired by the success of convolutional neural networks
(CNN) in many computer vision problems, deep CNN architectures \cite{Ejaz_cvpr2015,xiaogang_wang_cvpr2016,deepreid,joint_learning_cvpr16,hailin_shi,gated_siamese_eccv2016,de_cheng_2016}
have been widely used for person re-id. Using a deep model, the tasks
of feature representation learning and distance metric learning are
tackled jointly in a single end-to-end model. The state-of-the-art
re-id models are mostly based on deep learning; deep re-id is thus
the focus of this paper.

\begin{figure}
\begin{centering}
\subfigure[The cognitive process a human may take to match people]{\label{fig:a}
\includegraphics[scale=0.40]{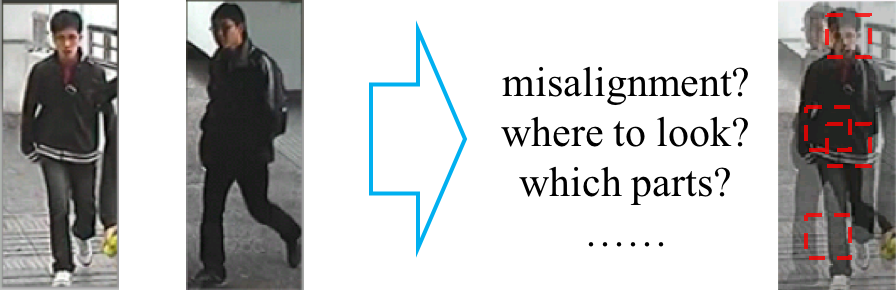} } \subfigure[Our model aims to imitate the human cognitive process]{\label{fig:b}
\includegraphics[scale=0.28]{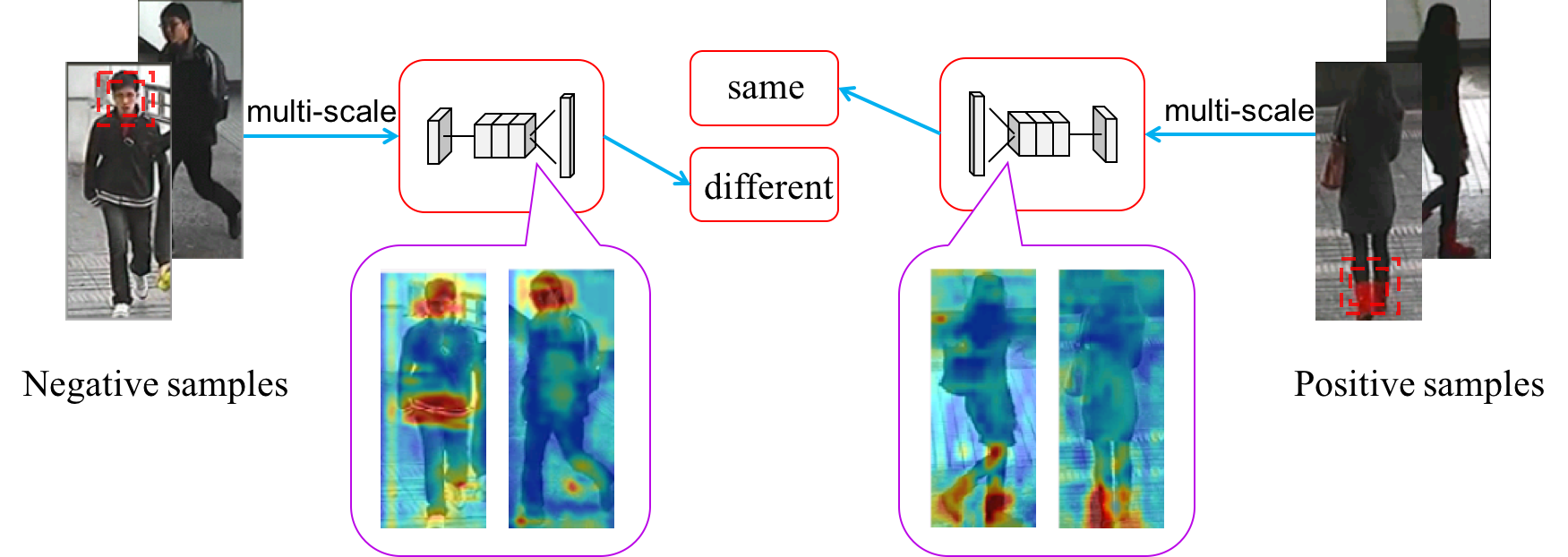} } 
\par\end{centering}
\caption{\label{fig:introduction}Multi-scale learning is adopted by  our MuDeep
to learn discriminative features at different spatial scales
and locations. }
\end{figure}

Learning discriminative feature representation is the key objective
of a deep re-id model. These features need to be computed at multiple
scales. More specifically, some people can be easily distinguished
by some global features such as gender and body build, whilst for
some others, detecting local images patches corresponding to, say
a handbag of a particular color or the type of shoes, would be critical
for distinguishing two otherwise very similarly-looking people. The
optimal matching results are thus only obtainable when features at
different scales are computed and combined. Such a multi-scale matching
process is likely also adopted by most humans when it comes to re-id.
In particular, humans typically compare two images from coarse to
fine. Taking the two images in Fig. \ref{fig:introduction}(a) as
an example. At the coarse level, the color and textual information
of clothes are very similar; humans would thus go down to finer scales
to notice the subtle local differences (\eg the hairstyle, shoe,
and white stripes on the jacket of the person on the left) to reach
the conclusion that these are two different people.

However, most existing re-id models compute features at a single scale
and ignore the factor that people are often only distinguishable at
the right spatial locations and scales. Existing models typically
adopt multi-branch deep convolutional neural networks (CNNs). Each
domain has a corresponding branch which consists of multiple convolutional/pooling
layers followed by fully connected (FC) layers. The final FC layer
is used as input to pairwise verification or triplet ranking losses
to learn a joint embedding space where people's appearance from different
camera views can be compared. However, recent efforts \cite{DBLP:journals/corr/GatysEB15,Mahendran15}
on visualizing what each layer of a CNN actually learns reveal that
higher-layers of the network capture more abstract semantic concepts
at global scales with less spatial information. When it reaches the
FC layers, the information at finer and local scales has been lost
and cannot be recovered. This means that the existing deep re-id architectures
are unsuitable for the multi-scale person matching.

In this work, we propose a novel multi-scale deep learning model (MuDeep) for re-id which aims to learn discriminative feature representations
at multiple scales with automatically determined scale weighting for
combining them (see Fig.~\ref{fig:introduction}(b)). More specifically,
our MuDeep network architecture is based on a Siamese network but
critically has the ability to learn features at different scales and evaluating their importance for cross-camera matching.
This is achieved by introducing two novel layers: \emph{multi-scale
stream layers} that extract images features by analyzing the person
images in multi-scale; and \emph{saliency-based learning fusion layer,
}which selectively learns to fuse the data streams of multi-scale and
generate the more discriminative features of each branch in MuDeep.
The multi-scale data can implicitly serve as a way of augmenting the
training data. In addition to the verification loss used by many previous
deep re-id models, we introduce a pair of classification losses at
the middle layers of our network, in order to strongly supervise multi-scale
features learning.

\section{Related Work}

\noindent \textbf{Deep re-id models} \quad{}Various deep learning
architectures have been proposed to either address visual variances
of pose and viewpoint \cite{deepreid},  learn better relative distances
of triplet training samples \cite{gua_gang_deep_reid}, or learn better
similarity metrics of any pairs \cite{Ejaz_cvpr2015}. To have enough
training samples, \cite{xiaogang_wang_cvpr2016} built upon inception
module a single deep network and is trained on multiple datasets; to
address the specific person re-id task, the neural network will be
adapted to a single dataset by a domain guided dropout algorithm. More
recently, an extension of the siamese network has been studied for person
re-id \cite{gated_siamese_eccv2016}. Pairwise and triplet comparison
objectives have been utilized to combine several sub-networks to form
a network for person re-id in \cite{joint_learning_cvpr16}. Similarly,
\cite{de_cheng_2016} employed triplet loss to integrate multi-channel
parts-based CNN models. 
To resolve the problem of large variations,
\cite{hailin_shi} proposed a moderate positive sample mining method
to train CNN. However, none of the models developed is capable of multi-scale feature computation as our model.

More specifically, the proposed deep re-id model differs from related
existing models in several aspects. (1) our MuDeep generalizes the
convolutional layers with multi-scale strategy and proposed multi-scale
stream layers and saliency-based learning fusion layer, which is
different from the ideas of combing multiple sub-networks \cite{joint_learning_cvpr16}
or channels \cite{de_cheng_2016} with pairwise or triplet loss. (2)
Comparing with \cite{xiaogang_wang_cvpr2016}, our MuDeep are simplified,
refined and flexible enough to be trained from scratch on either large-scale
dataset (\eg CUHK03) or medium-sized dataset (\eg CUHK01). Our experiments
show that without using any extra data, the performance of our MuDeep
is $12.41\%/4.27\%$ higher than that of \cite{xiaogang_wang_cvpr2016}
on CUHK01/CUHK03 dataset. (3) We improve the architecture of \cite{Ejaz_cvpr2015}
by introducing two novel layers to implement multi-scale and saliency-based
learning mechanisms. Our experiment results validate that the novel
layers lead to much  better performance  than \cite{Ejaz_cvpr2015}.

\vspace{0.02in}

\noindent \textbf{Multi-scale re-id } \quad{}The idea of multi-scale
learning for re-id was first exploited in \cite{Li_2015_ICCV}. However,
the definition of scale is different: It was defined as  different levels
of resolution rather than the global-to-local supporting region as in ours. Therefore, despite  similarity between terminology,
very different problems are tackled in these two works. The only multi-scale
deep re-id work that we are aware of is \cite{Liu:2016:MTC:2964284.2967209}.
Compared to our model, the model in \cite{Liu:2016:MTC:2964284.2967209}
is rather primitive and naive: Different down-sampled versions of
the input image are fed into shallower sub-networks to extract features
at different resolution and scale. These sub-networks are  combined
with a deeper main network for feature fusion. With an explicit network
for each scale, this network becomes computationally very expensive.
In addition, no scale weighting can be learned automatically and no
spatial importance of features can be modeled as in ours.

\begin{figure*}
\centering{}\includegraphics[scale=0.25]{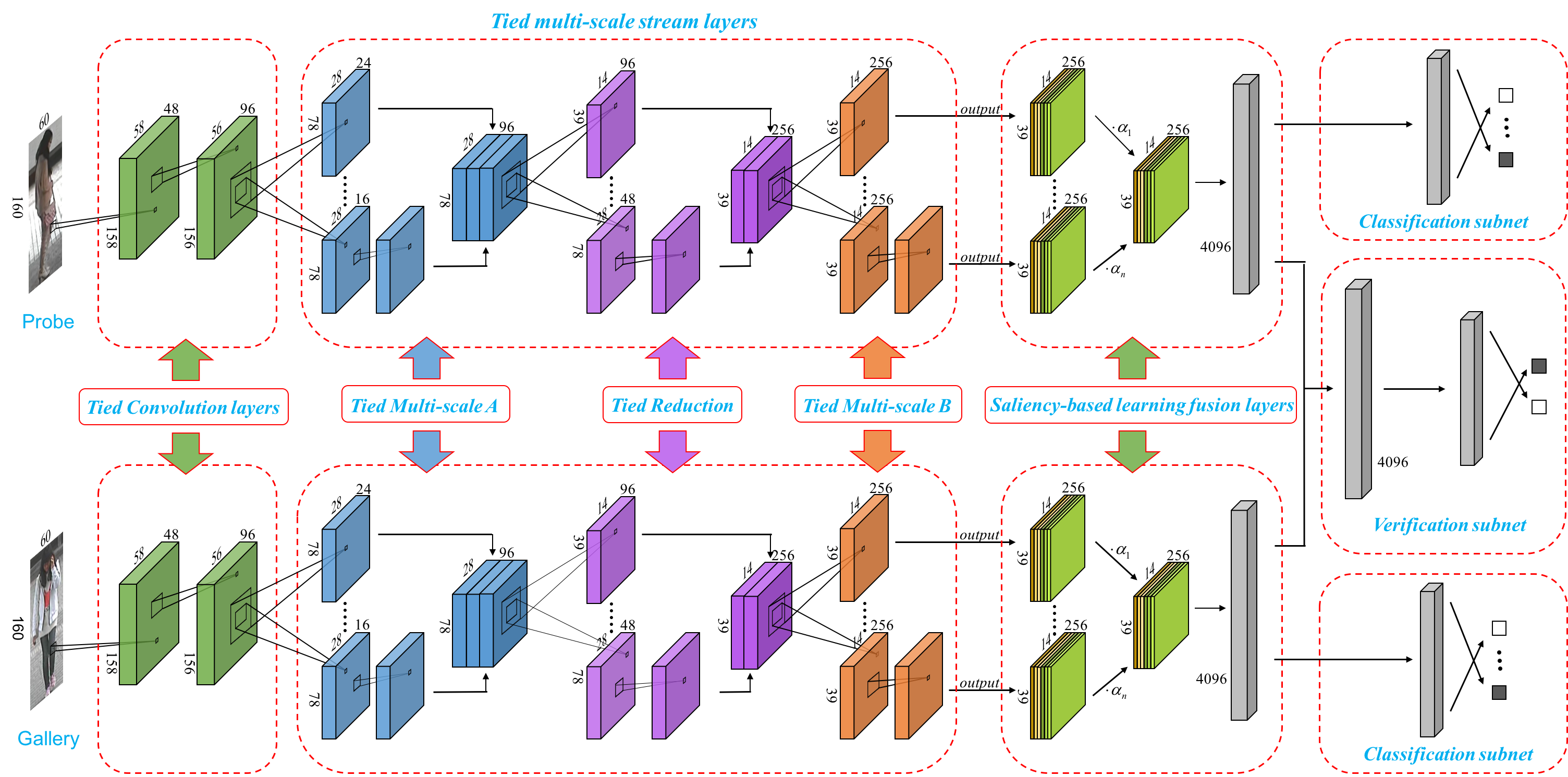}\caption{\label{fig:Overview}Overview of MuDeep architecture.}
\end{figure*}

\begin{table*}
\begin{centering}
\begin{tabular}{c||c|c||c}
\hline 
{\small{}{}Layers }  & \textcolor{black}{\small{}{}Stream id}{\small{}{} }  & \textcolor{black}{\small{}{}number@size}{\small{}{} }  & \textcolor{black}{\small{}output}\tabularnewline
\hline 
\multirow{4}{*}{\textcolor{black}{\small{}Multi-scale-A}} & \textcolor{black}{\small{}{}1}{\small{}{} }  & \textcolor{black}{\small{}{}1@$3\times3\times96$ AF \textendash 24@$1\times1\times96$
CF }{\small{} } & \multirow{4}{*}{\textcolor{black}{\small{}$78\times28\times96$}}\tabularnewline
\cline{2-3} 
 & \textcolor{black}{\small{}{}2}{\small{}{} }  & \textcolor{black}{\small{}{}24@$1\times1\times96$ CF}{\small{} } & \tabularnewline
\cline{2-3} 
 & \textcolor{black}{\small{}{}3}{\small{}{} }  & \textcolor{black}{\small{}{}16@$1\times1\times96$ CF \textendash{}
24@$3\times3\times96$ CF}{\small{}{} }  & \tabularnewline
\cline{2-3} 
 & \textcolor{black}{\small{}{}4}{\small{}{} }  & \textcolor{black}{\small{}{}16@$1\times1\times96$ CF \textendash 24@$3\times3\times96$
}{\small{}{}CF}\textcolor{black}{\small{}{}\textendash{} 24@$3\times3\times24$
CF}{\small{} } & \tabularnewline
\hline 
\multirow{3}{*}{\textcolor{black}{\small{}Reduction} } & \textcolor{black}{\small{}1}{\small{} }  & \textcolor{black}{\small{}{}1@$3\times3\times96$ MF{*}}{\small{}{}
}  & \multirow{3}{*}{\textcolor{black}{\small{}$39\times14\times256$}}\tabularnewline
\cline{2-3} 
 & \textcolor{black}{\small{}{}2}{\small{}{} } & \textcolor{black}{\small{}{}96@$3\times3\times96$ CF{*}}{\small{}{}
}  & \tabularnewline
\cline{2-3} 
 & \textcolor{black}{\small{}{}3}{\small{}{} } & \textcolor{black}{\small{}{}48@$1\times1\times96$ CF \textendash{}
56@$3\times3\times48$ CF\textendash{} 64@$3\times3\times56$ CF{*}}{\small{}{}
}  & \tabularnewline
\hline 
\multirow{5}{*}{\textcolor{black}{\small{}Multi-scale-B}} & \textcolor{black}{\small{}1}{\small{} }  & \textcolor{black}{\small{}{}256@$1\times1\times256$ CF}{\small{} } & \textcolor{black}{\small{}$39\times14\times256$}\tabularnewline
\cline{2-4} 
 & \textcolor{black}{\small{}{}2}{\small{}{} } & \textcolor{black}{\small{}{}64@$1\times1\times256$ CF \textendash 128@$1\times3\times64$
CF\textendash 256@$3\times1\times128$ CF}{\small{} } & \textcolor{black}{\small{}$39\times14\times256$}\tabularnewline
\cline{2-4} 
 & \multirow{2}{*}{\textcolor{black}{\small{}3}{\small{} }} & \textcolor{black}{\small{}{}64@$1\times1\times256$ CF \textendash{}
64@$1\times3\times64$ CF}{\small{} } & \multirow{2}{*}{\textcolor{black}{\small{}$39\times14\times256$}}\tabularnewline
 &  & \textcolor{black}{\small{}{}\textendash 128@$3\times1\times64$CF\textendash{}
128@$1\times3\times128$}{\small{}{} CF}\textcolor{black}{\small{}{}\textendash{}
256@$3\times1\times128$ CF}{\small{} } & \tabularnewline
\cline{2-4} 
 & \textcolor{black}{\small{}{}4}{\small{}{} }  & \textcolor{black}{\small{}{}1@$3\times3\times256$ AF{*} \textendash 256@$1\times1\times256$
CF}{\small{}{} }  & \textcolor{black}{\small{}{}$39\times14\times256$}\tabularnewline
\hline 
\end{tabular}
\par\end{centering}

{\small{}{}\caption{\label{tab:The-details-parameters-multi-stream}The parameters of
\textcolor{black}{tied multi-scale stream layers of MuDeep. }\textcolor{black}{\small{}{}{}{}{}Note
that (1) number@size indicates the number and the size of filters.
(2) {*} means the stride of corresponding filters is 2; the stride
of other filters is 1. We add 1 padding to the side of input data
stream if the corresponding side of C-filters is 3. (3) CF, AF, MF
indicate the C-filters, A-filters and M-filters respectively. A-filter
is the average pooling filter.}}
} 
\end{table*}

\noindent \textbf{Deep saliency modelling } \quad{}Visual saliency
has been studied extensively \cite{itti2001visual_attention,itti2001combination}.
It is typically defined in a bottom-up process. In contrast, attention mechanism
\cite{nature_attention} works in a top-down way and allows for salient
features to dynamically come to the front as needed. Recently, deep soft
attention modeling has received increasing interest as a means to
attend to/focus on local salient regions for computing deep features
\cite{sermanet2014attention,mnih2014recurrent,yang2016stacked,shuicheng_tip2016}. 
In this work, we use saliency-based learning strategy in a saliency-based
learning fusion layer to exploit both visual saliency
and attention mechanism. Specifically, with the multi-scale stream
layers, the saliency features of multiple scales are computed in multi-channel
(\eg in a bottom-up way); and a per channel weighting layer is introduced
to automatically discover the most discriminative feature channels
with their associated scale and locations. Comparing with \cite{shuicheng_tip2016} which adopts a spatial attention model, our model is much compact and can be learned from scratch on a small re-id dataset. When comparing the two models, our model, despite being much smaller, yields overall slightly better performance: on CUHK-01 dataset ours is $8\%$ lower than that of \cite{shuicheng_tip2016} but on the more challenging CUHK-03(detected) we got around $10\%$ improvement over that of \cite{shuicheng_tip2016}. Such a simple saliency learning architecture is shown to be very effective in our experiments.

\noindent \textbf{Our contributions} are as follows: (1) A novel multi-scale
representation learning architecture is introduced into the deep learning
architectures for person re-id tasks. (2) We propose a saliency-based
learning fusion layer which can learn to weight important scales in
the data streams in a saliency-based learning strategy. We evaluate
our model on a number of benchmark datasets, and the experiments show
that our models can outperform state-of-the-art deep re-id models,
often by a significant margin.

\section{Multi-scale Deep Architecture (MuDeep)}

\noindent \textbf{\textcolor{black}{Problem Definition. }}Typically,
person re-id is formulated only as a verification task \cite{larry_davis,XQDA,MFA,zhang_cooccurrence}.
In contrast, this paper formulates person re-id into two tasks: classification
\cite{wacv2016_enhanced,xiao2016endtoend} and verification \cite{Ejaz_cvpr2015,xiaogang_wang_cvpr2016}.
Specifically, given a pair of person images, our framework will categorize
them (1) either as the ``same person'' or ``different persons''
class, and (2) predict the person's identity.

\noindent \textbf{\textcolor{black}{Architecture Overview.}} As shown
in Fig.~\ref{fig:Overview}, MuDeep has two branches to process
each of image pairs. It consists of five components:\textcolor{red}{{}
}\emph{tied convolutional layer}s,\emph{ multi-scale stream layers
}(Sec. \ref{subsec:Tied-multi-scale-stream})\emph{, saliency-based
learning fusion layer }(Sec. \ref{subsec:Tied-saliency-based-fusion})\emph{,
verification subnet }and \emph{classification subnet }(Sec. \ref{subsec:Subnets-for-person})\emph{.}
Note that after each convolutional layer or fully connected layer,
batch normalization \cite{batch_normalization} is used before the
ReLU activation.

\noindent \textbf{Preprocessing by convolutional layers.} The input
pairs are firstly pre-processed by two convolutional layers with the
filters (C-filters) of 48@$3\times3\times3$ and 96@$3\times3\times48$;
furthermore, the generated feature maps are fed into a max-pooling
layer with filter size (M-filter) as 1@$3\times3\times96$ to reduce
both length and width by half. The weights of these layers are tied
across two branches, in order to enforce the filters to learn the
visual patterns shared by both branches.

\subsection{Multi-scale stream layers\label{subsec:Tied-multi-scale-stream}}

We propose multi-scale stream layers to analyze data streams in
multi-scale. The multi-scale data can implicitly serve as a way of augmenting
the training data. Different from the standard Inception structure
\cite{inception_v3}, all of these layers share weights between the
corresponding stream of two branches; however, within each two data
streams of the same branch, the parameters are not tied. The parameters
of these layers are shown in Tab. \ref{tab:The-details-parameters-multi-stream};
and please refer to  Supplementary Material for the visualization
of these layers.

\noindent \textbf{Multi-scale-A layer} analyses the data stream with
the size $1\times1$, $3\times3$ and $5\times5$ of the receptive field.
Furthermore, in order to increase both depth and width of this layer,
we split the filter size of $5\times5$ into two $3\times3$ streams
cascaded (\ie stream-4 and stream-3 in Table \ref{tab:The-details-parameters-multi-stream}).
The weights of each stream are also tied with the corresponding stream
in another branch. Such a design is in general inspired by, and
yet different from  Inception architectures \cite{inception_v1,inception_v3,inception_v4}.
The key difference lies in the factors that  the weights  are not tied between
any two streams from the same branch, but are tied between  two
corresponding streams of different branches.

\noindent \textbf{Reduction} \textbf{layer} further passes the data
streams in multi-scale, and halves the width and height of feature
maps, which should be, in principle, reduced from $78\times28$ to
$39\times14$. We thus employ Reduction layer to \emph{gradually}
decrease the size of feature representations as illustrated in Table
\ref{tab:The-details-parameters-multi-stream}, in order to avoid
representation bottlenecks. Here we follow the design principle of
``avoid representational bottlenecks'' \cite{inception_v3}. In
contrast to directly use max-pooling layer for decreasing feature
map size, our ablation study shows that the Reduction layer, if replaced
by max-pooling layer, will leads to more than $10\%$ absolute points
lower than the reported results of Rank-1 accuracy on the CUHK01 dataset
\cite{cuhk01}. Again, the weights of each filter here are  tied
for paired streams.

\noindent \textbf{Multi-scale-B} \textbf{layer} serves as the last
stage of high-level features extraction for the multiple scales of
$1\times1$, $3\times3$ and $5\times5$. Besides splitting the $5\times5$
stream into two $3\times3$ streams cascaded (\ie stream-4 and stream-3
in Table \ref{tab:The-details-parameters-multi-stream}). We can further
decompose the $3\times3$ C-filters into one $1\times3$ C-filter
followed by $3\times1$ C-filter \cite{inception_v4}. This leads
to several benefits, including reducing the computation cost on $3\times3$
C-filters, further increasing the depth of this component, and being capable of
 extracting asymmetric features from the receptive field. We still
tie the weights of each filter.

\subsection{Saliency-based learning fusion layer \label{subsec:Tied-saliency-based-fusion}}

This layer is proposed to fuse the outputs of multi-scale stream layers.
Intuitively, with the output processed by previous layers, the resulting
data channels have redundant information: Some channels may capture
relative important information of persons, whilst others may only
model the background context. The saliency-based learning strategy
is thus utilized here to automatically discover and emphasize the
channels that had extracted highly discriminative patterns, such as
the information of head, body, arms, clothing, bags and so on, as illustrated in Fig. \ref{fig:Attention-Map-of}. Thus,
we assume $\mathbf{F}_{i\star}$ represents the input feature maps
of $i-$th stream $(1\leq i\leq4)$ in each branch and $\mathbf{F}_{ij}$
represents the $j-$th channel of $\mathbf{F}_{i\star}$, \ie $(1\leq j\leq256)$
and $\mathbf{F}_{ij}\in\mathbb{R}^{39\times14}$. The output feature
maps denoted as $\mathbf{G}$ will fuse the four streams; $\mathbf{G}_{j}$
represents the $j-$th channel map of $\mathbf{G}$, which is computed
by: 
\begin{equation}
\mathbf{G}_{j}=\sum_{i=1}^{4}\mathbf{F}_{ij}\cdot\alpha_{ij}~~~(1\leq j\leq256)\label{eq: attention_fusion}
\end{equation}

\noindent where $\alpha_{ij}$ is the scalar for $j-$th channel of
$\mathbf{F}_{i\star}$; and the saliency-weighted vector $\alpha_{i\star}$
is learned to account for the importance of each channel of stream $\mathbf{F}_{i\star}$;
$\alpha_{i\star}$ is also tied.

A fully connected layer is appended after saliency-based learning
fusion layer, which extracts features of 4096-dimensions
of each image. The idea of this design is 1) to concentrate the saliency-based
learned features and reduce dimensions, and 2) to increase the efficiency
of testing.

\subsection{Subnets for person Re-id \label{subsec:Subnets-for-person}}

\noindent \textbf{Verification subnet}\quad{}accepts feature pairs
extracted by previous layers as input, and calculate distance with
\emph{feature difference layer}, which followed by a fully connected
layer of 512 neurons with 2 softmax outputs. The output indicates
the probability of \char`\"{}same person\char`\"{} or \char`\"{}different
persons\char`\"{}. Feature difference layer is employed here to fuse
the features of two branches and compute the distance between two
images. We denote the output features of two branches as $\mathbf{G}^{1}$
and $\mathbf{G}^{2}$ respectively. The feature difference layer computes
the difference $\mathbf{D}$ as $\mathbf{D}=\left[\mathbf{G}^{1}-\mathbf{G}^{2}\right].*\left[\mathbf{G}^{1}-\mathbf{G}^{2}\right]$.
Note that (1) `$.*$' indicates the element-wise multiplication;
the idea behind using element-wise subtraction is that if an input image pair is labelled
\char`\"{}same person\char`\"{}, the features generated by multi-scale
stream layers and saliency-based learning fusion layers should be
similar; in other words, the output values of feature difference layer should be close to zero; otherwise, the values have different responses.
(2) We empirically compare the performance of two difference layer operations including $\left[\mathbf{G}^{1}-\mathbf{G}^{2}\right].*\left[\mathbf{G}^{1}-\mathbf{G}^{2}\right]$
and $\left[\mathbf{G}^{1}-\mathbf{G}^{2}\right]$. Our experiment shows
that the former achieves $2.2\%$ higher performance than the latter
on Rank-1 accuracy on CUHK01. 

\noindent \textbf{Classification subnet} \quad{}In order to learn
strong discriminative features for appearance representation, we add
classification subnet following saliency-based learning fusion layers
of each branch. The classification subnet is learned to classify images
with different pedestrian identities. After extracting 4096-D features in
saliency-based learning fusion layers, a softmax with \emph{N} output
neurons are connected, where \emph{N} denotes the number of pedestrian
identities.

\section{Experiments\label{experiment}}

\subsection{Datasets and settings}

\noindent \textbf{Datasets. }The proposed method is evaluated on three
widely used datasets, \ie CUHK03 \cite{deepreid}, CUHK01 \cite{cuhk01}
and VIPeR \cite{viper}. The CUHK03 dataset includes $14,096$ images
of $1,467$ pedestrians, captured by six  camera views. Each person has $4.8$ images on average. 
Two types of person images are  provided \cite{deepreid}: manually
labelled pedestrian bounding boxes (labelled) and bounding boxes automatically
detected by the deformable-part-model detector \cite{felzenszwalb2010dt_part}
(detected). The manually labelled images generally are of higher quality
than those detected images. We use the settings of both manually \emph{labelled}
and automatically \emph{detected} person images on the standard splits
in \cite{deepreid} and report the results in Sec. \ref{subsec:Experiment-on-CUHK03-Detected}
and Sec. \ref{subsec:Experiment-on-CUHK03-Labelled} respectively.
CUHK01 dataset has 971 identities with 2 images per person of each
camera view. As in \cite{cuhk01}, we use as probe the images from
camera A and take those from camera B as gallery. Out of all data,
we select randomly 100 identities as the test set. The remaining identities
for training and validation. The experiments are repeated over 10
trials. For all the experiments, we train our models from the scratch.
VIPeR has $632$ pedestrian pairs in two views with only one image
per person of each view. We split the dataset and half of pedestrian
pairs for training and the left for testing as in \cite{Ejaz_cvpr2015}
over 10 trials. In addition, we also evaluate proposed method on two video-based re-id datasets, \textit{i.e.},  iLIDS-VID dataset \cite{wang2014person} and PRID-2011 dataset \cite{hirzer2011person}.  The iLIDS-VID dataset contains 300 persons, which are captured by two non-overlapping cameras. The sequences range in length from 23 to 192 frames, with an average number of 73. The PRID-2011 dataset contains 385 persons for camera view A; 749 persons for camera view B, with sequences lengths of 5 to 675 frames. These two camera views have no nonoverlapping. Since the primary focus of this paper is on image-based person re-id, we employ the simplest feature fusion scheme for video re-id:  Given a video sequence, we compute features of each frame which are aggregated by max-pooling to form video level representation. In contrast, most of the state-of-the-art video-based re-id methods \cite{mclaughlin2016recurrent,liu2015spatio,wang2014person,karanam2015sparse,li2015multi,karanam2015person} utilized the RNN models such as LSTM to perform temporal/sequence video feature fusion from each frame.


\vspace{0.03in}

\noindent \textbf{Experimental settings.} On the CUHK03 dataset, in term
of training set used, we introduce two specific experimental settings;
and we report the results for both settings: (a)\textbf{ Jointly}:
as in \cite{xiaogang_wang_cvpr2016}, under this setting the model is firstly
trained with the image set of both labelled and detected CUHK03 images,
and for each task, the corresponding image set is used to fine-tune
the pre-trained networks. (b) \textbf{Exclusively}: for each of the ``labelled''
and ``detected'' tasks, we only use the training data from each
task without using the training data of the other task. 

\vspace{0.03in}

\noindent \textbf{Implementation details.} We implement our model
based on the Caffe framework \cite{caffe} and we make our own implementation
for the proposed layers. We follow the training strategy used in \cite{Ejaz_cvpr2015}
to first train the network without classification subnets; secondly,
we add the classification subnets and freeze other weights to learn
better initialization of the identity classifier; finally we train classification
loss and verification loss simultaneously, with a higher loss weight
of the former. The training data include positive and negative pedestrian
pairs. We augment the data to increase the training set size by 5 times
with the method of random 2D translation as  in \cite{deepreid}.
The negative pairs are randomly sampled as twice the  number of positive
pairs. We use the stochastic gradient descent algorithm with the mini-batch
size of 32. The learning rate is set as $0.001$, and gradually decreased
by $\nicefrac{1}{10}$ every 50000 iterations. The size of input image
pairs is\footnote{To make a fair comparision with \cite{Ejaz_cvpr2015}, the
 input images are resized to $60\times160\times3$. } $60\times160\times3$. Unless specified otherwise, the dropout ratio
is set as $0.3$. The proposed MuDeep get converged in $9\sim12$ hours on re-id dataset
on a NVIDIA TITANX GPU. Our MuDeep needs around 7GB GPU memory. Code and
models will be made available on the first author's webpage.

\vspace{0.03in}

\noindent \textbf{Competitors. }We compare with the deep learning
based methods including DeepReID \cite{deepreid}, Imp-Deep \cite{Ejaz_cvpr2015},
En-Deep \cite{wacv2016_enhanced}, and G-Dropout \cite{xiaogang_wang_cvpr2016},
Gated\_Sia \cite{gated_siamese_eccv2016}, EMD \cite{hailin_shi}, SI-CI \cite{joint_learning_cvpr16}, and MSTC\footnote{We re-implement \cite{Liu:2016:MTC:2964284.2967209} for evaluation purpose.} \cite{Liu:2016:MTC:2964284.2967209}, as well as other non-deep
competitors, such as Mid-Filter \cite{mid_rui_zhao}, and XQDA \cite{XQDA},
LADF \cite{LADF}, eSDC \cite{unsupervised_per_reid}, LMNN \cite{LMNN},
and LDM \cite{face_metric}.

\vspace{0.03in}

\noindent \textbf{Evaluation metrics.} In term of standard evaluation
metrics, we report the Rank-1, Rank-5 and Rank-10 accuracy with single-shot
setting in our paper. For more detailed results using \emph{Cumulative Matching
Characteristics} (CMC) curves, please refer to the Supplementary Material.

\subsection{Results on CUHK03-Detected\label{subsec:Experiment-on-CUHK03-Detected}}

On the CUHK03-Detected dataset, our results are compared with the state-of-the-art
methods in Table \ref{tab:Results-of-CUHK03.}. 

Firstly and most importantly, our best results \textendash{} MuDeep
(jointly) outperforms all the other baselines at all ranks. Particularly, we notice that our results
are significantly better than both the methods of using hand-crafted features
and the recent deep learned models. This validates the efficacy of
our architectures  and suggests that the proposed multi-scale and saliency-based
learning fusion layer can help extract  discriminative features
 for person re-id. 

Secondly, comparing with the Gated\_Sia \cite{gated_siamese_eccv2016}
which is an extension of Siamese network with the gating function
to selectively emphasize fine common local patterns from data, our
result is $7.54\%$ higher at Rank-1 accuracy. This suggests that
our framework can better analyze the multi-scale patterns from data
than Gated\_Sia \cite{gated_siamese_eccv2016}, again thanks to the
novel  multi-scale stream layers and saliency-based learning
fusion layers.

Finally, we  further compare our results on both ``Jointly''
and ``Exclusively'' settings. The key difference is that in the
``jointly'' settings, the models are also trained with the data
of CUHK03-Labelled, \ie the images with manually labelled pedestrian
bounding boxes. As explained in \cite{deepreid}, the quality of labelled
images is generally better than those of detected images. Thus with
more data of higher quality used for training, our model under the ``Jointly'' setting
can indeed beat our model under the ``Exclusively'' setting. However, the
improved margins between these two settings at  all Ranks
are very small if compared with the margin between our results and
the results of the other methods. This suggests that our multi-scale
stream layers have efficiently explored and augmented the training
data; and with such multi-scale information explored, we can better  train our model. Thanks to our multi-scale stream layers,
our MuDeep can still achieve good results with less training data,
\emph{i.e.} under the Exclusively setting.

\begin{table}
\centering{}%
\begin{tabular}{c||ccc}
\hline 
Dataset  & \multicolumn{3}{c}{``Detected''}\tabularnewline
\hline 
Rank  & 1  & 5  & 10\tabularnewline
\hline 
SDALF\cite{farenzena2010reidentify_symmetry}  & 4.87  & 21.17  & 35.06\tabularnewline
eSDC \cite{unsupervised_per_reid}  & 7.68  & 21.86  & 34.96\tabularnewline
LMNN \cite{LMNN}  & 6.25  & 18.68  & 29.07\tabularnewline
XQDA \cite{XQDA}  & 46.25  & 78.90  & 88.55\tabularnewline
LDM \cite{face_metric}  & 10.92  & 32.25  & 48.78\tabularnewline
\hline 
DeepReid \cite{deepreid}  & 19.89  & 50.00  & 64.00\tabularnewline
MSTC \cite{Liu:2016:MTC:2964284.2967209} & 55.01 & \textendash{} & \textendash{}\tabularnewline
Imp-Deep \cite{Ejaz_cvpr2015}  & 44.96  & 76.01  & 83.47\tabularnewline
SI-CI \cite{joint_learning_cvpr16}  & 52.17  & 84.30  & 92.30\tabularnewline
Gated\_Sia \cite{gated_siamese_eccv2016}  & 68.10  & 88.10  & 94.60\tabularnewline
EMD \cite{hailin_shi}  & 52.09  & 82.87  & 91.78\tabularnewline
\hline 
MuDeep (Jointly)  & \textbf{75.64  } & \textbf{	94.36  	} & \textbf{97.46}\tabularnewline
MuDeep (Exclusively)  & 75.34   & 94.31  & 97.40\tabularnewline
\hline 
\end{tabular}\caption{\label{tab:Results-of-CUHK03.}Results of CUHK03-Detected dataset.
\textcolor{red}{}}
\end{table}

\subsection{Results on CUHK03-Labelled \label{subsec:Experiment-on-CUHK03-Labelled}}

The results of CUHK03-Labelled dataset are shown
in Table \ref{tab:Results-of-CUHK03-Labelled} and we can make the following
observations.

Firstly, in this setting, our MuDeep still outperforms the other competitors
by clear margins on all the rank accuracies. Our result is $4.27\%$
higher than the second best one \textendash{} G-Dropout\cite{xiaogang_wang_cvpr2016}.
Note that the G-Dropout adopted the domain guided dropout strategies
and it utilized much more training data in this task. This further
validates that our multi-scale stream layers can augment the training
data to exploit more information from medium-scale dataset rather
than scaling up the size of training data; and saliency-based learning
fusion layers can better fuse the output of multi-scale information
which can be used for person re-id.

Secondly, we can draw a similar conclusion as from the CUHK03-Detected results:
Our MuDeep with ``Jointly'' setting is only marginally better than
that with ``Exclusively'' setting. This means that more available
related training data can always help improve the performance of deep
learning model; and this also validates that our multi-scale stream
layers and saliency-based fusion layers can help extract and fuse
the multi-scale information and thus cope well with less training data. 

\begin{table}
\begin{centering}
\begin{tabular}{c||ccc}
\hline 
Dataset  & \multicolumn{3}{c}{``Labelled'' }\tabularnewline
\hline 
Rank  & 1  & 5  & 10 \tabularnewline
\hline 
SDALF\cite{farenzena2010reidentify_symmetry}  & 5.60  & 23.45  & 36.09 \tabularnewline
eSDC \cite{unsupervised_per_reid}  & 8.76  & 24.07  & 38.28 \tabularnewline
LMNN \cite{LMNN}  & 7.29  & 21.00  & 32.06 \tabularnewline
XQDA \cite{XQDA}  & 52.20  & 82.23  & 92.14 \tabularnewline
LDM \cite{face_metric}  & 13.51  & 40.73  & 52.13 \tabularnewline
\hline 
DeepReid \cite{deepreid}  & 20.65  & 51.50  & 66.50 \tabularnewline
Imp-Deep \cite{Ejaz_cvpr2015}  & 54.74  & 86.50  & 93.88 \tabularnewline
G-Dropout\cite{xiaogang_wang_cvpr2016}  & 72.60  & $92.30${*}  & 94.30{*} \tabularnewline
EMD \cite{hailin_shi}  & 61.32  & 88.90  & 96.44 \tabularnewline
\hline 
MuDeep (Jointly)  & \textbf{76.87 }  & \textbf{96.12 }  & \textbf{98.41 }\tabularnewline
MuDeep (Exclusively)  & 76.34   & 95.96  & 98.40\tabularnewline
\hline 
\end{tabular}
\par\end{centering}
\caption{\label{tab:Results-of-CUHK03-Labelled}Results of CUHK03-Labelled
dataset. Note that: {*} represents the results reproduced from \cite{xiaogang_wang_cvpr2016}
with the model trained only by CUHK03 dataset. }
\end{table}

\subsection{Results on CUHK01 and VIPeR}

\noindent \noindent \textbf{CUHK01 dataset}. Our MuDeep is trained
only on CUHK01 dataset without using extra dataset. As listed in Table
\ref{tab:Results-of-CUHK01}, our approach obtains $79.01\%$ on Rank-1
accuracy, which can beat all the state-of-the-art; and is $7.21\%$
higher than the second best method \cite{joint_learning_cvpr16}.
This further shows the advantages of our framework.

\vspace{0.03in}

\noindent \noindent \textbf{VIPeR dataset.} This dataset is extremely challenging
due to small data size and low resolution. In particular, this dataset has relatively
small number of distinct identities and thus the positive pairs for
each identity are much less if compared with the other two datasets. Thus
for this dataset, our network is initialized by the model pre-trained
on CUHK03-Labelled dataset with the ``Jointly'' setting. The training
set of VIPeR dataset is used to fine-tune the pre-trained network.
We still use the same network structure except changing the number
of neurons on the last layer in the classification subnet. The results
on VIPeR dataset are listed in Table \ref{tab:Results-of-VIPER}. The
results of our MuDeep remains competitive and outperforms all compared methods.
\vspace{0.03in}

\begin{figure}
\begin{centering}
\includegraphics[scale=0.45]{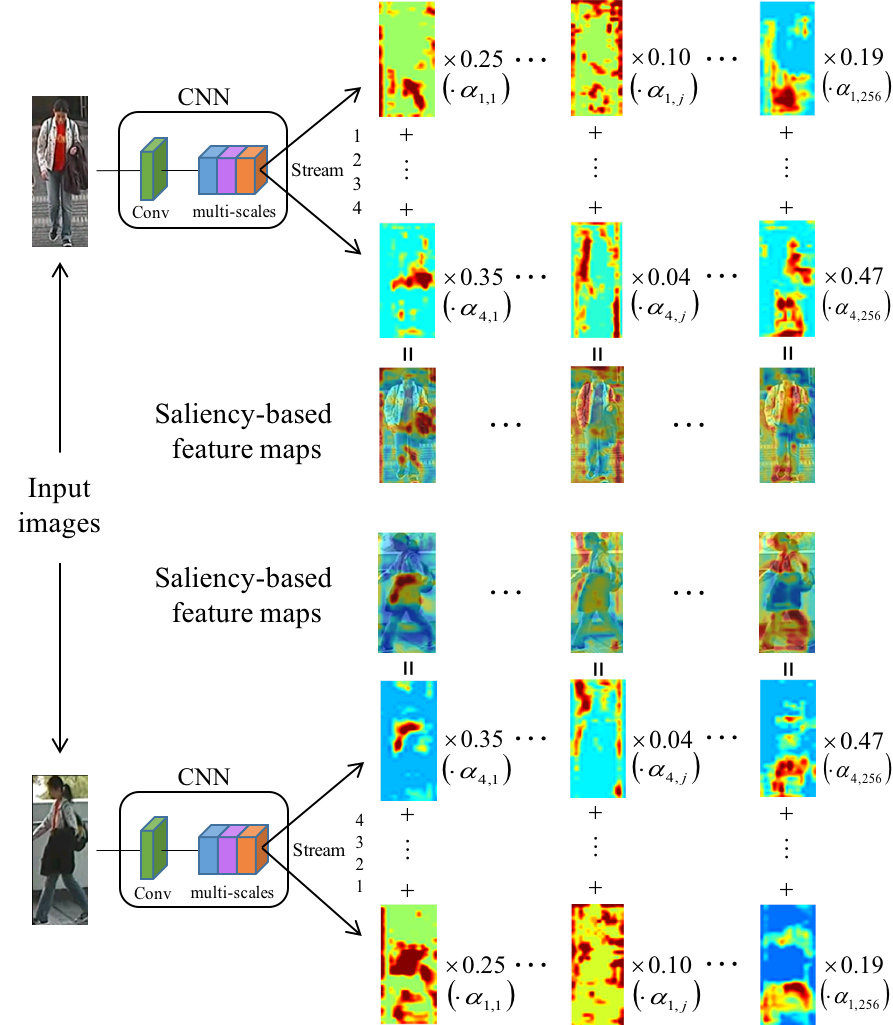} 
\par\end{centering}
\caption{\label{fig:Attention-Map-of}Saliency Map of $\mathbf{G}$ in Eq (\ref{eq: attention_fusion}).}
\end{figure}

\noindent \noindent \textbf{Qualitative visualization.} We give some
qualitative results of visualizing the saliency-based learning maps
on CUHK01. The visualization of our saliency-based learning
maps is computed from saliency-based learning fusion layer in Fig.
\ref{fig:Attention-Map-of}. Given one input pair of images (left
of Fig. \ref{fig:Attention-Map-of}), for each branch, the saliency-based
learning fusion layer combines four data streams into a single data
stream with selectively learning learned from the saliency of each
data stream. The heatmaps of three channels are visualized for each
stream and each branch is shown on the right side of Fig. \ref{fig:Attention-Map-of}.
Each row corresponds to one data stream; and each column is for
one channel of heatmap of features. The weight $\alpha$ (in Eq.~(\ref{eq: attention_fusion}))
for each feature channel is learned and updated accordingly with the
iteration of the whole network. The three channels illustrated in
Fig. \ref{fig:Attention-Map-of} have a high reaction (i.e. high values)
in the heatmaps on discriminative local patterns of the input image pair.
For example, the first and third columns highlight the difference
of clothes and body due to the existence of different visual patterns
learned by our multi-scale stream layers. Thus these two channels
have relative higher $\alpha$ weights, whilst the second column
models the background patterns which are less discriminative for the
task of person re-id and results in lower $\alpha$ value. Our saliency-based
learning fusion layer can automatically learn  the optimal $\alpha$
weights from training data.

\begin{table}
\centering{}%
\begin{tabular}{c||ccc}
\hline 
Rank  & 1  & 5  & 10\tabularnewline
\hline 
KISSME \cite{kCCA,KISSME}  & 29.40  & 60.18  & 74.44\tabularnewline
SDALF\cite{farenzena2010reidentify_symmetry}  & 9.90  & 41.21  & 56.00\tabularnewline
eSDC \cite{unsupervised_per_reid}  & 22.84  & 43.89  & 57.67\tabularnewline
LMNN \cite{LMNN}  & 21.17  & 49.67  & 62.47\tabularnewline
LDM \cite{face_metric}  & 26.45  & 57.68  & 72.04\tabularnewline
\hline 
DeepReid \cite{deepreid}  & 27.87  & 58.20  & 73.46\tabularnewline
G-Dropout \cite{xiaogang_wang_cvpr2016}  & 66.60  & \textendash{}  & \textendash{}\tabularnewline
MSTC \cite{Liu:2016:MTC:2964284.2967209} & 64.12 & \textendash{} & \textendash{}\tabularnewline
Imp-Deep \cite{Ejaz_cvpr2015}  & 65.00  & 88.70  & 93.12\tabularnewline
SI-CI \cite{joint_learning_cvpr16}  & 71.80  & 91.35{*}  & 95.23{*}\tabularnewline
EMD \cite{hailin_shi}  & 69.38  & 91.03  & 96.84\tabularnewline
\hline 
MuDeep  & \textbf{79.01}  & \textbf{97.00}  & \textbf{98.96}\tabularnewline
\hline 
\end{tabular}\caption{\label{tab:Results-of-CUHK01}Results of CUHK01 dataset. {*}: reported
from CMC curves in \cite{joint_learning_cvpr16}.}
\end{table}

\begin{table}
\centering{}%
\begin{tabular}{c||ccc}
\hline 
Rank  & 1  & 5  & 10\tabularnewline
\hline 
kCCA\cite{kCCA}  & 30.16  & 62.69  & 76.04\tabularnewline
Mid-Filter \cite{mid_rui_zhao}  & 29.11  & 52.34  & 65.95\tabularnewline
RPLM \cite{RPLM}  & 27.00  & 55.30  & 69.00\tabularnewline
MtMCML \cite{MtMCML}  & 28.83  & 59.34  & 75.82\tabularnewline
LADF \cite{LADF}  & 30.22  & 64.70  & 78.92\tabularnewline
XQDA \cite{XQDA}  &   40.00    &  68.13    &   80.51\tabularnewline 
\hline 
Imp-Deep \cite{Ejaz_cvpr2015} & 34.81  & 63.61  & 75.63\tabularnewline
G-Dropout \cite{xiaogang_wang_cvpr2016} & 37.70  & \textendash{}  & \textendash{}\tabularnewline
MSTC \cite{Liu:2016:MTC:2964284.2967209} & 31.24 & \textendash{} & \textendash{}\tabularnewline
SI-CI \cite{joint_learning_cvpr16} & 35.76 & 67.40 & 83.50\tabularnewline
Gated\_Sia \cite{gated_siamese_eccv2016} & 37.90 & 66.90 & 76.30\tabularnewline
EMD \cite{hailin_shi} & 40.91 & 67.41 & 79.11\tabularnewline
\hline 
MuDeep  & \textbf{43.03} & \textbf{74.36} & \textbf{85.76}\tabularnewline
\hline 
\end{tabular}\caption{\label{tab:Results-of-VIPER}Results on the VIPeR dataset}
\end{table}

\subsection{Ablation study\label{subsec:multiscale-ablation}}

\noindent \textbf{Multi-scale stream layers.} We  compare
our multi-scale stream layers with three variants of Inception-v4 \cite{inception_v4} on CUHK01 dataset.
Specifically, Inception-v4 has Inception A and Inception B modules,
both of which are compared against here. Furthermore, we also compare Inception A+B
structure which is constructed  by connecting Inception A, Reduction, and Inception
B modules. The Inception A+B structure is the most similar one to our
multi-scale stream layers except that (1) we modify some parameters; (2) the weights of our layers are tied between the corresponding streams of two branches. Such weight
tieing strategy enforces each paired stream of our two branches to extract
the common patterns. With the input of each image pair, we use
Inception A, Inception B, and Inception A+B as the based network structure
to predict whether this pair is the ``same person'' or ``different
persons''. The results are compared in Table \ref{tab:inception}.
We can see that our MuDeep architecture has the best performance over the other baselines. This
shows that our MuDeep is the most powerful at learning discriminative
patterns than the Inceptions variants, since the multi-scale stream
layers can more effectively extract multi-scale information and the saliency-based
learning fusion layer facilitates the automatic selection of the
important feature channels. 

\vspace{0.02in}
\noindent \textbf{Saliency-based learning fusion layer and classification
subset. }To further investigate the contributions of the fusion layer and the 
classification subset, we compare three variants of our model with one of or both of the two components removed  on CUHK01 dataset. In Table \ref{tab:alpha}, the ``\textendash{}
Fusion'' denotes our MuDeep without using the  fusion layer; and ``\textendash ClassNet''
indicates our MuDeep without the classification subnet; and ``\textendash{}
Fusion \textendash{} ClasNet'' means that MuDeep has neither fusion
layer nor classification subnet. The results in Table
\ref{tab:alpha} show that our full model has the best performance
over the three variants. We thus conclude that both components help and the combination of the two can further boost the performance. 

\subsection{Further evaluations\label{subsec:further-evaluations}}

\noindent \textbf{Multi-scale vs.~Multi-resolution.} Due to the often different camera-to-object distances and the resultant object image resolutions, multi-resolution re-id is also interesting on its own and could potentially complement multi-scale re-id. Here a  simplest multi-resolution multi-scale re-id model is formulated, by training the proposed multi-scale models at different resolutions followed by fusing the different model outputs. We consider the original resolution ($60\times160$) and a lower one ($45\times145$). The feature fusion is done by concatenation. We found that the model trained at the lower resolution achieves lower results and when the two models are fused, the final performance is around $1-2\%$ lower than the model learned at the original resolution alone. Possible reasons include (1) All three datasets have images of similar resolutions; and (2) more sophisticated multi-resolution learning model is required which can automatically determine the optimal resolution for measuring similarity given a pair of probe/gallery images.

\noindent  \textbf{Results on  video-based re-id}. Our method can be evaluated on  iLIDS-VID and PRID-2011 datasets. Particularly,  two datasets are randomly split into $50\%$ of persons for training and $50\%$ of persons for testing.  We follow the evaluation protocol in \cite{hirzer2011person} on PRID-2011 dataset and only consider  the first 200 persons who appear in both cameras.   We compared our model with the results reported in \cite{hirzer2011person, wang2014person, mclaughlin2016recurrent}.
The results are listed in Table \ref{tab:Results-of-VIDEO}. These results are higher than those in \cite{hirzer2011person, wang2014person}, but lower than those in \cite{mclaughlin2016recurrent} (only slightly on PRID-2011)\footnote{We also note that our results are better than those of most video-based re-id specialist models listed in Table 1 of \cite{mclaughlin2016recurrent}.}.  These results are  quite encouraging and we expect that if the model is extended to a CNN-RNN model, better performance can be achieved.

\begin{table}
\begin{centering}
\begin{tabular}{c||ccc}
\hline 
Rank & 1 & 5 & 10\tabularnewline
\hline 
Inception A & 60.11 & 85.30 & 92.44\tabularnewline
Inception B & 67.31 & 92.71 & 97.43\tabularnewline
Inception A+B & 72.11 & 91.90 & 96.45\tabularnewline
\hline 
MuDeep & \textbf{79.01} & \textbf{97.00} & \textbf{98.96}\tabularnewline
\hline 
\end{tabular}
\par\end{centering}
\caption{\label{tab:inception}Results of comparing with different inception
models on the CUHK01 dataset. }

\end{table}

\begin{table}
\begin{centering}
\begin{tabular}{c||ccc}
\hline 
Rank & 1 & 5 & 10\tabularnewline
\hline 
\textendash{} Fusion  & 77.88 & 96.81 & 98.21\tabularnewline
\textendash{} ClassNet & 76.21 & 94.47 & 98.41\tabularnewline
\textendash{} Fusion \textendash{} ClasNet & 74.21 & 92.10 & 97.63\tabularnewline
\hline 
MuDeep & \textbf{79.01} & \textbf{97.00} & \textbf{98.96}\tabularnewline
\hline 
\end{tabular}
\par\end{centering}
\caption{\label{tab:alpha}Results of comparing with the variants of MuDeep
on the CUHK01 dataset. Note that ``\textendash Fusion'' means that saliency-based
learning fusion layer is not used; ``\textendash ClassNet'' indicates
that the classification subnet is not used in the corresponding structure.}
\end{table}

\begin{table}
\centering{}%
\begin{tabular}{c||ccc||ccc}
\hline 
Dataset & \multicolumn{3}{c||}{PRID-2011} & \multicolumn{3}{c}{iLIDS-VID} \tabularnewline
\hline 
Rank  & 1  & 5  & 10 & 1  & 5  & 10 \tabularnewline
\hline 
RCNvrid\cite{mclaughlin2016recurrent}  & \textbf{70}  & \textbf{90}  & \textbf{95} & \textbf{58}  & \textbf{84}  & \textbf{91} \tabularnewline
STA \cite{liu2015spatio}  & 64  & 87  & 90 & 44  & 72  & 84 \tabularnewline
VR \cite{wang2014person}  & 42  & 65  & 78 & 35  & 57  & 68 \tabularnewline
SRID \cite{karanam2015sparse}  & 35  & 59  & 70 & 25  & 45  & 56 \tabularnewline
AFDA \cite{li2015multi}  & 43  & 73  & 85 & 38  & 63  & 73 \tabularnewline
DTDL \cite{karanam2015person}  & 41  & 70  & 78 & 26  & 48  & 57 \tabularnewline 
DDC \cite{hirzer2011person} & 28  & 48  & 55 & \textendash{}  & \textendash{} & \textendash{} \tabularnewline 
\hline 
MuDeep  & 65  & 87  & 93 & 41  & 70  & 83\tabularnewline
\hline 
\end{tabular}\caption{\label{tab:Results-of-VIDEO}Results on the PRID-2011 and iLIDS-VID datasets}
\end{table}

\section{Conclusion }

We have identified the limitations of existing deep re-id
models in the lack of multi-scale discriminative feature learning.
To overcome the limitation, we have presented a novel deep architecture
\textendash{} MuDeep to exploit the multi-scale and saliency-based
learning strategies for re-id. Our model has achieved state-of-the-art
performance on several benchmark datasets.

\vspace{0.03in}
\noindent \textbf{Acknowledgments.}  This work was supported in part by two NSFC projects ($\#U1611461$ and $\#U1509206$) and one project from STCSM ($\#16JC1420401$).

{\small{}\bibliographystyle{ieee}
\bibliography{egbib}
 }{\small \par}
\end{document}